\pdfoutput=1
\documentclass{article} 
\usepackage{colm2024_conference}

\usepackage{microtype}
\usepackage{hyperref}
\usepackage{url}
\usepackage{booktabs}
\definecolor{darkblue}{rgb}{0, 0, 0.5}
\hypersetup{colorlinks=true, citecolor=darkblue, linkcolor=darkblue, urlcolor=darkblue}
\usepackage{graphicx}
\usepackage{subcaption}
\usepackage{multirow, multicol}
\usepackage{diagbox}
\usepackage{algorithm}
\usepackage{algpseudocode}
\usepackage{comment}

\usepackage{tabularx}
\usepackage{xspace}
\usepackage{amsmath}
\usepackage{float}
\usepackage{xcolor}
\usepackage{tablefootnote}

\usepackage{pifont}
\usepackage{amssymb}

\usepackage{enumitem}

\newcommand{\TODO}[1]{{\color{red}TODO: #1}}
\newcommand{\commentAH}[1]{{\color{blue}AH: #1}}

\newcommand{\commentGC}[1]{{\color{blue}GC: #1}}

\newcommand{\wilbur}{\textsc{Wilbur}\xspace}

\title{\wilbur: Adaptive In-Context Learning for Robust \newline and Accurate Web Agents}

\author{Michael Lutz, Arth Bohra \thanks{Equal contribution. Work done while at Bardeen.} \\
Department of Eletrical Engineering and Computer Science\\
University of California Berkeley\\
Berkeley, CA, USA \\
\texttt{\{michaeljlutz,arthbohra\}@berkeley.edu} \\
\And
Manvel Saroyan, Artem Harutyunyan, Giovanni Campagna \\
Bardeen, Inc. \\
San Francisco, CA, USA \\
\texttt{\{manvel,artem,giovanni\}@bardeen.ai}
}

\colmpreprint
\begin{document}

\maketitle
\begin{abstract}
In the realm of web agent research, achieving both generalization and accuracy remains a challenging problem. Due to high variance in website structure, existing approaches often fail. Moreover, existing fine-tuning and in-context learning techniques fail to generalize across multiple websites. We introduce \wilbur, an approach that uses a differentiable ranking model and a novel instruction synthesis technique to optimally populate a black-box large language model’s prompt with task demonstrations from previous runs. To maximize end-to-end success rates, we also propose an intelligent backtracking mechanism that learns and recovers from its mistakes. Finally, we show that our ranking model can be trained on data from a generative auto-curriculum which samples representative goals from an LLM, runs the agent, and automatically evaluates it, with no manual annotation. \wilbur achieves state-of-the-art results on the WebVoyager benchmark, beating text-only models by 8\% overall, and up to 36\% on certain websites. On the same benchmark, \wilbur is within 5\% of a strong multi-modal model despite only receiving textual inputs, and further analysis reveals a substantial number of failures are due to engineering challenges of operating the web.
\end{abstract}


\section{Introduction}

The rise of large language models has led to various attempts at creating intelligent agents that interact with the web through a browser, also known as \textit{web agents} \citep{gur2022understanding, kagaya2024rap, kim2024language}. By encoding the Document Object Model (DOM) and optionally a screenshot of the page in a multimodal fashion, state-of-the-art web agents have obtained noteworthy success rates on various tasks on the web \citep{he2024webvoyager, zheng2024gpt4vision}. Yet, the success rate of even the best web agents is a far cry from that of experienced people familiar with the websites and tasks at hand.

We hypothesize that the disparity in success rate is not only due to the reasoning limitations of the underlying LLM, but also due to the need to learn how specific websites works. Even for a person, it is not enough to know how to operate the web: instead, faced with a never-seen-before website, one needs to explore, try different approaches, and adjust. Only after succeeding at the task once (or a few times), one can perform the task without hitting dead ends or clicking the wrong link. At the same time, there are more than a billion websites in the world \citep{number-of-websites}. It is implausible that any LLM can memorize all of them just from pretraining, thus any zero-shot approach is likely to fail \citep{kim2024language}.

\begin{figure*}[t]
  \centering
  \includegraphics[width=\linewidth]{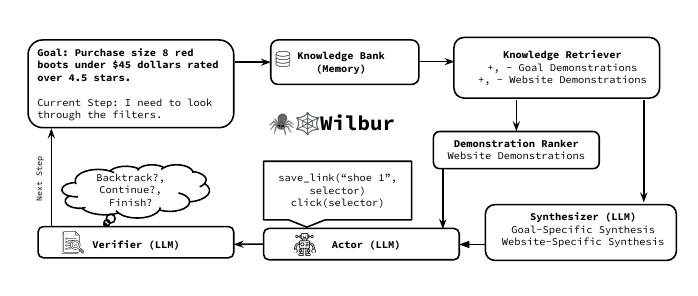}
  \caption{The \wilbur Agent, which utilizes retrieval, synthesis, action, and verification steps to accomplish tasks on the web.}
  \label{fig:wilbur-methodology}
\end{figure*}

To address this challenge and improve the generality of web agents, we propose \wilbur, an agent with two novel capabilities (Fig.~\ref{fig:wilbur-methodology}):
\begin{itemize}
\item \textit{explore, reflect, and backtrack}: faced with a novel website, \wilbur proceeds by executing an action sampled from an LLM. After observing the new page state, it queries a reflection LM to verify that the action contributed progress toward the goal. If verification fails, \wilbur dynamically \textit{backtracks} to a previous successful state, while storing the failure in the model's context for all future steps.
\item \textit{retrieve demonstrations from a scalable knowledge bank}: We include both \textit{goal-conditioned} demonstrations, which teach \wilbur how to perform a similar task on a potentially unseen website, and \textit{website-conditioned} demonstrations, which teach \wilbur how to act on a similar web page, regardless of the overall task. These two sources of knowledge are complementary and help \wilbur generalize.
\end{itemize}



As the limited LLM context window can only fit a small number of demonstrations, we train a dedicated \textit{demonstration ranking model} to select the most helpful ones. This model is trained to predict whether the actions will lead to a successful execution or not and optimally populates a model's context. Additionally, following \citet{bohra-etal-2023-byoc}, we propose to summarize a large sample of successful and unsuccessful actions into concise instructions. This combination of explicit examples and summarized instructions allows the model to see a few details while also gather insight from many runs, including unsuccessful ones.

Finally, in order to quickly acquire knowledge of new websites and new tasks, we propose an autocurriculum which generates plausible goals to populate demonstration banks \citep{clark2003bootstrapping, mcclosky-etal-2006-effective, wang2023voyager}. Applying an LLM-based automatic scoring step to evaluate an agent's execution, our approach quickly populates a dataset of trajectories, both successful and unsuccessful. These executions can be fed back into the agent in the future, through task demonstrations and instruction synthesis, to further improve success rate. 

To evaluate our approach, we have \wilbur on the WebVoyager~\citep{he2024webvoyager} benchmark, where we achieve a new text-only state-of-the-art result of 53\%, 8\% higher than the previous state of the art, and within 5\% of a strong multimodal model. Compared to a strong baseline with retrying but no backtracking, using backtracking and loop detection improves by 6\%. Furthermore, our approach of adaptive in-context-learning and autocurriculum leads to an additional 12\% improvement, without any annotated data, which strongly shows the importance of learning the websites where the agent is applied.

\subsection{Contributions}

The contributions of this paper are as follows:
\begin{itemize}
    \item We propose the first web agent that is able to recover from delayed mistakes, by modeling the web agent task as graph exploration over the web and adding the ability to navigate back to a previous state in the graph.

    \item To reduce the number of backtracks and improve accuracy, we propose learning in-context from previous executions of the agent, querying on similar pages and goals. These executions are both selected as task demonstrations using a novel trained model, and summarized into succinct instructions, allowing a large number of executions to be used while limiting the prompt size.

    \item To bootstrap the agent on new websites and task, we are the first to apply an autocurriculum strategy with an LLM scoring step to the web agent task, allowing us to obtain high quality training data without human feedback.

    \item We evaluated our approach end-to-end on the WebVoyager benchmark, and we find our agent outperforms the zero-shot baseline by 18\%, and the text-only state of the art by 8\%.
    
\end{itemize}

\section{Related Work}
\paragraph{Web Agents}
The field of web agents has recently seen an increase in research interest, due to the general availability of powerful LLMs. These agents are tested on a variety of benchmarks, some highly specialized \citep{NEURIPS2022_82ad13ec} and others with general goals on a diverse set of websites \citep{shi2017world, liu2018reinforcement, he2024webvoyager, deng2023mind2web}. 

Common state of the art approaches for web agents include splitting the agent's work amongst an actor, a retriever, a planner, and a verifier or critique step \citep{gur2022understanding, kagaya2024rap, he2024webvoyager}. Previous work also proposed using a code-generation step, with LLM feedback to critique the generated code \citep{gur2023real,sun2024adaplanner}.  Critically, these works only include the ability to retry failures, and do not backtrack to previous steps, so they cannot recover mistakes that are not immediately apparent. \citet{ma2023laser} recognized the need for backtracking, but used a fixed state-space specific for the WebShop benchmark, instead of operating directly on the agent's trajectory on the web.

Due to context window limitations, many previous works did not use few-shot task demonstrations, preferring zero-shot prompting techniques such as ReAct \citep{yao2022react} or fine-tuning dedicated models \citep{furuta2023multimodal, xu2021grounding}. \citet{sridhar2023hierarchical} and \citet{deng2023mind2web} proposed summarizing the current web page to reduce the prompt size, but did not include previous demonstrations in the summary. \cite{kim2024language} included task demonstrations in the LLM prompt, but only used few positive experiences related to the specific subtask.
\wilbur is the first agent to summarize positive and negative task demonstrations, and separate both website-conditioned and goal-conditioned examples.


\paragraph{In-Context and Few-shot Learning}
Beyond web agents, in-context learning, also known as few-shot learning, has been applied to a number of tasks \citep{NEURIPS2020_1457c0d6, gao-etal-2021-making, hendrycks2021measuring, min2022rethinking}. In ICL, demonstrations taken from a training set are included as examples in the to the prompt of an LLM, optionally with explanations \citep{lampinen-etal-2022-language}. Raw task demonstrations are limited by the size of the context window. 
Previous work by \citet{bohra-etal-2023-byoc} proposed summarizing training examples into succinct instructions. They only evaluated on classification, and did not use negative examples. Additionally, existing approaches either use a fixed set of few-shot training examples, or naively include the closest examples by embedding similarity. We propose instead to use a dedicated model to predict the suitability of each specific example.

\section{The \wilbur Agent}
\label{sec:methodology}

\subsection{Problem Statement}
The web agent challenge can be modeled under a Partially Observable Markov Decision Process (POMDP). The agent receives a natural language goal $g$ which requires a multi-step execution where each action $a \in \mathcal{A}$ modifies the web page (clicking, typing, etc.) and/or extracts textual information for the user.

The agent's state space $\mathcal{S}$ represents a combination of previously extracted text and the website environment's true state, itself comprised of current DOM, network requests, state of the website's backend, etc. Because the backend is not accessible to a web agent, the agent's observation space $\mathcal{O}$ is limited to all visible DOM elements as well as the text extracted by the agent so far. Specifically, we treat $o \in \mathcal{O}$ as follows:
$$
o = (\text{URL}, \text{DOM}, \ \text{Extracted Text})
$$

In $o$, the DOM is formatted as text. The specific format is described in Appendix~\ref{sec:dom-formatting}.

Provided the current $o_t$ and the goal $g$, the agent takes an action according to a policy $\pi$:
$$\pi(o_t,g) \rightarrow a_t, \ \text{where} \ o \in \mathcal{O}, \ a \in \mathcal{A}$$

resulting in a trajectory $\tau = \{(o_0,a_0),...,(o_t,a_t)\}$ at timestep $t$. When the agent finishes execution, the reward $r$ can be calculated deterministically by a benchmark or estimated by a self-evaluation module $\hat{r}(\tau,g,\textit{answer}) \in [0,1]$ run at the end of an execution, where $\textit{answer}$ is the textual output of the agent.

\subsection{\wilbur During Inference Time}
\begin{figure}[t]
\begin{minipage}{0.49\linewidth}
\begin{algorithm}[H]
\caption{\wilbur agent loop}
\label{algo:inference}
\small
\begin{algorithmic}[1]
\Require goal $g$, initial state $o$
\State $D_g^+, D_g^- \gets \text{RetrieveForGoal}(D_\tau, g)$
\State $l_g \gets \text{SynthesizeForGoal}(D_\tau^+, D_\tau^-, g)$
\State $\tau \gets \emptyset$
\State $\varphi \gets \emptyset$
\State $\textit{p} \gets g$
\State $\textit{done} \gets \textsc{continue}$
\While{$\textit{done} \ne \textsc{finish}$}
    \State $D_a^+, D_a^- \gets \text{RetrieveForAction}(D_a, p)$
    \State $l_a \gets \text{SynthesizeForAction}(D_a^+, D_a^-, g, o)$
    \State $a \gets \text{Actor}(o, \tau, \varphi, D_a^+, l_a, l_g)$
    \State $o', \varphi' \gets \text{Execute}(a)$
    \State $\tau' \gets \tau \cup \{ (a, o') \}$
    \State $\textit{done}, p, \varphi' \gets \text{Verify}(o, o', \tau', \varphi')$
    \State $\varphi \gets \varphi \cup \varphi'$
    \If{$\textit{done} = \textsc{continue}$}
        \State $o \gets o'$
        \State $\tau \gets \tau'$
    \ElsIf{$\textit{done} = \textsc{backtrack}$}
        \State $o_b, \tau, p \gets \text{Previous}(\tau)$
        \State $o \gets \text{Revert}(o_b)$
    \EndIf
\EndWhile
\State \textbf{return} $(\tau, \text{Answer}(o, \tau, g))$
\end{algorithmic}
\end{algorithm}
\end{minipage}
\hfill
\begin{minipage}{0.49\linewidth}
\begin{figure}[H]
\centering
\includegraphics[width=0.9\linewidth]{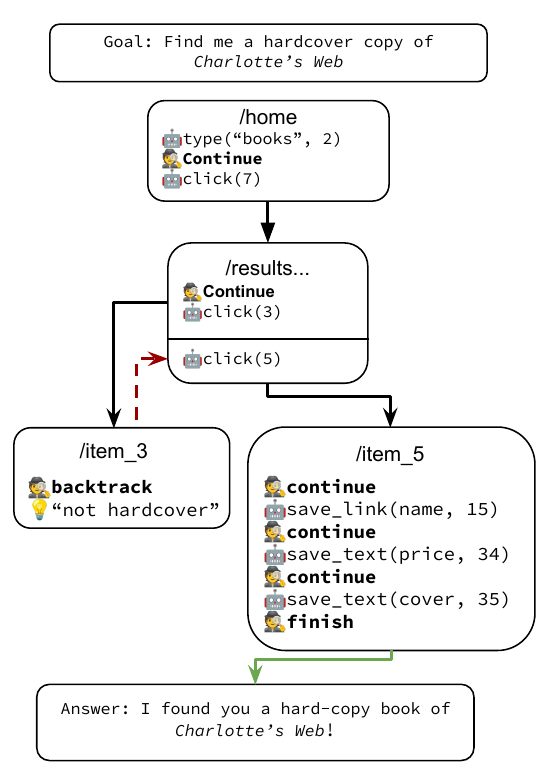}
\caption{An example of \wilbur backtracking to previous URL after failure.}
\label{fig:graph-traversal}
\end{figure}
\end{minipage}
\end{figure}

Given the goal and the current state of the page, the \wilbur repeatedly executes actions according to policy $\pi$ until the task is predicted to have finished or until backtracking is necessary. The formal algorithm is given in Algorithm~\ref{algo:inference}, and an example is shown in Fig.~\ref{fig:graph-traversal}.

At each step of the execution, \wilbur makes use of the following sub-modules:
\begin{enumerate}
    \item the \textit{demonstration retriever} queries a demonstration bank of full-length trajectories $\mathcal{D}_\tau$ and finds the relevant ones $D_\tau \subset \mathcal{D}_\tau$; individual action demonstrations $D_a$ are also queried from $\mathcal{D}_a$; the retriever obtains both positive (successful) demonstrations $D^+$ and negative (unsuccessful) demonstrations $D^-$.
    \item the \textit{knowledge synthesizer} summarizes the demonstrations into a description of learnings $l$; $D_\tau$ and $D_a$ are summarized separately.
    \item the \textit{actor} references $D_\tau$, $D_a$, and $l$ to predict an action $a$, given the current state $o$, next step plan $p$ form previous step, and feedback $\varphi$ if returning from a backtrack.
    \item the \textit{executor} performs action $a$ on the website and obtains the new observable state $o'$ as well as execution feedback $\varphi$.
    \item the \textit{reflection} module compares $o$ and $o'$ before determining whether to backtrack, continue, or finish; if backtrack $\rightarrow$ update $\varphi$; if continue $\rightarrow$ plan next step $\textit{p'}$
    \item at the end of an execution, the \textit{answer module} produces the textual response required by the goal using the final observable state $o$, and agent's trajectory $\tau$
\end{enumerate}

In the rest of this section, we describe each \wilbur module in detail.

\paragraph{Demonstration Retrieval}
The goal of the demonstration retriever is to identify previous relevant trajectories and actions to use as guidance. Full-length trajectories provide information relevant to planning, while action demonstrations help the actor correctly interact with website elements. Hence, we store two types of demonstration banks: a bank $\mathcal{D}_{\tau}$ of entire trajectories and a bank $D_a$ of isolated actions relevant to the current step action plan. 

\wilbur queries $\mathcal{D}_{\tau}$ with the goal $g$ and queries $\mathcal{D}_a$ with the current step $p$, using cosine similarity between an embedding of the query and an embedding of the trajectory or action in the bank. The $k_g$ goal-conditioned trajectories $D_g$ are then split into $D_g^+$ (successful) and $D_g^-$ (unsuccessful) demonstrations.

We include negative examples to help \wilbur avoid previously discovered pitfalls. For example, through negative goal-conditioned trajectory demonstrations, \wilbur can learn how specific steps can lead to downstream failure of tasks. Similarly, showing the agent negative website-conditioned tasks teaches it how to avoid mistakes given the specific nuances of certain webpages (e.g. saving the wrong text, clicking the wrong element).

\paragraph{Action Demonstration Reranking} In order to have action demonstrations scale across multiple websites and types of pages (e.g. search pages, documentation, etc.), we must also factor in DOM similarity. Additionally, the quality of positive action demonstrations strongly impacts performance, and simple cosine similarity is not sufficient to determine whether a demonstration will actively help the actor. While a demonstration might be similar, there may exist slight differences in respective DOMs that lead the actor astray.

Hence, after retrieving the top $k_a$ action demonstrations $d \in D_a$, we re-rank them:
\begin{align*}
\vec{h}_{o_d}, \vec{h}_{p_d}, \vec{h}_{a_d}, \vec{h}_p, \vec{h}_o &= \text{Embedding}(o_d, p_d, a_d, p, o) \\
\textit{sim}_d &= \alpha_1 (\vec{h}_{o_d}^T \vec{h}_o) + \alpha_2 (\vec{h}_{p_d}^T\vec{h}_{p}) \\
\textit{score}_d &= \textit{sim}_d \times \text{MLP}(\vec{h}_{o_d} \| \vec{h}_{p_d} \| \vec{h}_{a_d} \| \vec{h}_{o} \| \vec{h}_{p}) \\
d^+ &\sim \text{Softmax}(\textit{score}_{d \in D_a})
\end{align*}

where where $\alpha_1$ and $\alpha_2$ are hyperparameters. The ranking model is an MLP that encodes the embeddings of the demonstrations (observation, plan, action) as well as the current observation and plan, and it is trained to predict whether a demonstration leads to a successful execution or not, as a 0-1 score.
After computing the score of each demonstrations and normalizing with softmax, we then sample $k_a^+$ successful executions to include in the actor's context.

\paragraph{Synthesizing Demonstrations}
We wish to retrieve a large space of relevant demonstrations, so the actor has access to a diverse set of previous experiences. Yet, we cannot include the raw $k$ demonstrations in the actor's prompt due to context window limitations. To overcome this, \wilbur calls a synthesizer LLM to distill the essence of multiple demonstrations into actionable insights.  This guides the actor in performing the next action by highlighting common patterns, strategies, or pitfalls identified across the demonstrations.

\paragraph{Action Prediction and Execution}

Given the goal $g$, $k' < k$ raw demonstrations, the synthesized learnings $l$, and the current observable state  of the webpage $o$, the actor predicts:
\begin{align*}
a = \pi(o, g) = \text{Actor}(o, \tau, \varphi, D_a^+, l_a, l_g)
\end{align*}

The actor is implemented as an LLM which produces executable code in a domain-specific language optimized for web actions. The DSL also includes operations to save content from the page in the state. The full definition of the agent DSL is given in Appendix~\ref{sec:dsl-definition}. If the actor does not produce valid DSL (according to its syntax or semantics), it is prompted again to attempt to produce a different $a$, adding the DSL compilation feedback to the context.

The DSL is executed by an interpreter in the web browser. If execution fails, for example because the selector does not match an interactable element on the page, the whole step is marked as failed, new feedback $\varphi$ is computed, and the agent backtracks. After execution, the new observed state $o'$ is computed based on the new URL and new content of the page. 

\paragraph{Reflection and Backtracking}
The purpose of the reflection LM is to assess the effectiveness of actions taken by \wilbur. After the actor executes an action, resulting in a new observation $o'$, the reflector checks whether the action has reasonably completed the planned step. It takes into consideration the previous observed state $o$, new observed state $o'$, the action $a$, the plan $p$, and the current goal to make its judgment:
$$
v, \varphi, p' = \text{Reflect}(o, o', a, p, g)
$$
where $v$ is a ternary verdict:
\begin{itemize}
    \item \textsc{finish}: the current goal was completed successfully and the agent is done
    \item \textsc{continue}: proceed by completing planned step $p'$ next
    \item \textsc{backtrack}: backtrack and try an alternative action with feedback $\varphi$
\end{itemize}

The reflector uses both a rule-based comparison algorithm that checks for differences between $o$ and $o'$, and an LLM to compute the verdict. Qualitatively, we observe that incorrect executions often result in no change in the DOM, for example because the agent tries to click on a button that is disabled because a form needs to be filled first. Hence, the rule-based comparison ensures that execution performed had the desired effect on the page. The additional reflection step then checks that the new state of the page corresponds to the expected state according to the plan.

If the agent backtracks, it returns to the most recent observation $o_{prev}$ that is possible to return to. Because the backend is real and not simulated, not all state changes can be reverted. In the current implementation, \wilbur returns to the most recent state that corresponded to a navigation (change in page URL). The new state is applied by that refreshing or navigating to that URL, which resets the DOM on the page.

Overall, the reflector helps in ensuring \wilbur remains on the most promising path towards the goal at every step by preventing wasted efforts on ineffective actions. 

\paragraph{Answering Model}
Once \wilbur finishes the execution, it leverages a final LLM call to deliver a human-readable answer as a response. Given the goal, execution history, and extracted text, it produces a summary of \wilbur's trajectory and addresses the initial goal.

\subsection{Learning websites with \wilbur}
In order to populate \wilbur's demonstration banks and train the knowledge model, we leverage a multi-step auto-curriculum to collect reference trajectories (Algorithm~\ref{algo:wilbur_training}):

\begin{enumerate}
    \item An auto-curriculum is run on a batch of websites and record predicted end-to-end success using an execution evaluation LM.
    \item A more challenging goal-generation process is run conditioned on initial goals and utilizing task and goal demonstrations from trajectories recorded in the first step.
    \item We train our knowledge model to predict success likelihood of actions in the follow-up run which reference demonstrations in the first run.
\end{enumerate}

\paragraph{Autocurriculum Goal Generation}
In order to model realistic use-cases and goals on the web, we model the goal generation process as a function of a website's DOM representation:
$$
G = \text{GenerateGoals}(w)
$$
where $G = \{g_1, g_2, ..., g_n\}$ is a batch of goals that are reasonably achievable given the starting state of a website, sampled from an LLM. We instruct the LLM to produce information extraction and DOM interaction goals in approximately equal proportion.

During the second phase of the auto-curriculum, we condition on previously generated goals in order to develop diverse and more challenging follow-up goals $G'$:
$$
G' = \text{GenerateGoals}(G, w)
$$

\paragraph{Self-Evaluation}
In order to evaluate an agent's execution on a goal from the auto-curriculum, we model self-evaluation $\hat{r} \in [0, 1]$ as a function of the agent's execution trajectory $\tau$ and returned text $\textit{answer}$. $\hat{r}$ is predicted by an LLM which evaluates the entire execution trajectory with regards to the goal's requirements.


\paragraph{Knowledge Model Training}
\label{km:architecture}
The first run of the auto-curriculum generates demonstrations queried during the second run of the auto-curriculum. As such, the follow-up run generates training data for the knowledge model. We train the knowledge model to predict action success $v_a$ as estimated in the reflection step, using binary cross-entropy loss.




\section{Evaluation}
We implemented \wilbur as a browser extension in a commercial web automation platform.
In this section, we evaluate it on the WebVoyager benchmark \citep{he2024webvoyager}. For reproducibility, upon publication, we will release detailed evaluation results, as well as the auto-curriculum training data.

\subsection{Experimental Setup}
\paragraph{Benchmark}
WebVoyager is a benchmark of 643 goals, divided across 15 diverse real websites. Tasks include navigation, information retrieval, transactions, general question-answering. The tasks are performed on the actual websites, not on simulators or sandbox environments. The score is obtained at the end of the task from automatic evaluation by GPT4V~\citep{openai2023gpt4}, which uses the screenshot of the page at the end of the task. Note though that \wilbur is a text-only model and does not use screenshots to predict actions.

We chose WebVoyager because it uses real websites and, unlike other benchmarks, the score only measures task success rate.  Agents are not penalized for following different trajectories than the reference one, as long as they complete the task and obtain the right answer.

\paragraph{Ablation Study}
To further study the effect of the different components of our methodology, we perform an ablation study. We compare  \wilbur against the following baselines:
\begin{itemize}
\item Zero-shot: a baseline with no backtracking capabilities and no task demonstrations.
\item $+$ Backtracking: a zero-shot agent with the ability to verify and backtrack.
\item $+$ Demonstrations: the agent is additionally prompted with positive-only task demonstrations, obtained from the auto-curriculum training data by embedding similarity with no dedicated model.
\item $+$ Synthesis: in addition to positive task demonstrations, we synthesize instructions from positive and negative demonstrations.
\end{itemize}

\paragraph{Hyperparameters}
For our experiments, we use OpenAI GPT-4 Turbo~\citep{openai2023gpt4} (\texttt{gpt-4-0125-preview}) as the underlying LLM. We set the temperature to 0 for all calls, except for the actor LLM during autocurriculum, which has a temperature of $0.4$. We use OpenAI's \textit{text-embedding-3-large} as the embedding model, compressed down to a dimension of (1536, 1) for vector storage. For every goal, we retrieve 20 goal-conditioned demonstrations by cosine-similarity, of which we use 2 negative and 3 positive for instruction synthesis. Similarly for every execution step, we retrieve 20 website-conditioned demonstrations by cosime-similar, from which we select 5 positive to include in the prompt, and we use 5 positive and 5 negative examples for instruction synthesis.

For the knowledge model, we use a three-layer fully-connected MLP, using a ReLU non-linearity between hidden layers and a sigmoid at the end. The input concatenates embedding vectors of demonstration DOM, demonstration plan, demonstration action, current DOM, and current plan for a total size of 7680. The intermediate layers have hidden size 200.

\paragraph{Autocurriculum Training}
We sample 254 goals during the initial phase of the autocurriculum across the websites used in WebVoyager. Of these, 72\% were marked successful during the self-evaluation phase. Overall, the knowledge bank includes 183 goals and 634 action steps across all websites.

To train the knowledge model, we record total action steps across the follow-up goal executions. Of all actions, approximately 61\% were marked as as successful during the autocurriculum. We split the action step demonstrations into 4319 demonstrations for training and 1080 for evaluation. On the held-out set, the knowledge model achieves an accuracy of 93.1\% and F1 score of 0.942 after training for 8 epochs with batch-size 32 and learning rate 0.001.

\subsection{Results}

\begin{table}
\caption{Evaluation of \wilbur on WebVoyager (automatic evaluation). ``text'' is a text-only model, ``multi'' is multimodal (text and vision). All \wilbur results are text-only.}
\label{table:webvoyager}
\scriptsize
\centering
\begin{tabular}{lrrrrrrrr}
\toprule
 & Allrecipes & Amazon & Apple & ArXiv & GitHub & Booking & ESPN & Coursera \\
\midrule
\citet{he2024webvoyager} (text) & 57.8 & 43.1 & 36.4 & 50.4 & \bf 63.4 & 2.3 & 38.6 & 24.6 \\
\hline
Zero-shot          & 40.0     & 41.5     & 48.9     & 39.5     & 22.0     & 4.5      & 31.8     & 31.0 \\
$+$ Backtrack      & 35.6     & 43.9     & 41.9     & 39.5     & 31.7     & 2.3      & 47.7     & \bf 73.3 \\
$+$ Demonstrations & 28.9     & 43.9     & 62.8     & \bf 62.8 & 17.1     & 18.3     & 40.1     & 61.9 \\
$+$ Synthesis      & 53.3     & \bf 48.8 & \bf 65.1 & 55.8     & 12.2      & 13.6     & 47.7     & 61.9 \\
\wilbur            & \bf 60.0 & 43.9     & 60.5     & 51.2     & 22.0     & \bf 38.6 & \bf 59.1 & 51.1  \\
\hline
\citet{he2024webvoyager} (multi) & 51.1 & 52.9 & 62.8 & 52.0 & 59.3 & 32.6 & 47.0 & 57.9 \\
\bottomrule
\toprule
& Dictionary & BBC News & Flights & Maps & Search & Huggingface & Wolfram & \bf Overall \\
\midrule
\citet{he2024webvoyager} (text) & 66.7 & 45.2 & \bf 7.1 &\bf 62.6 & \bf 75.2 & 31.0 & 60.2 & 44.3 \\
\hline
Zero-shot          & 46.5     & 54.8     & 2.4 & 41.5 & 55.8     & 41.9     & 15.2     & 34.4 \\
$+$ Backtrack      & 74.4     & 69.0     & 0.0 & 34.1 & 62.8     & 41.9     & 13.0     & 40.6 \\
$+$ Demonstrations & 72.1     & 76.2     & 2.4 & 43.9 & 67.4     & 53.5     & \bf 71.7 & 48.4 \\
$+$ Synthesis      & 76.7     & 66.7     & 0.0 & 46.3 & 72.1     & \bf 58.1 & 67.4     & 49.9 \\
\wilbur            & \bf 86.0 & \bf 81.0 & 0.0 & 39.0 & 67.4     & 53.5     & 65.2     & \bf 52.6 \\
\hline
\citet{he2024webvoyager} (multi) & 71.3 & 60.3 & 51.6 & 64.3 & 77.5 & 55.8 & 60.9 & 57.1 \\
\bottomrule
\end{tabular}
\end{table}

Results of running the agent are shown in Table~\ref{table:webvoyager}. On this benchmark, \wilbur outperforms the state-of-the-art text-only model by \citet{he2024webvoyager} by 8\%. Specifically, we observe that \wilbur outperforms the text-only state of the art on most websites except GitHub and the Google websites. It improves substantially on the very hard Booking.com case, from around 2-4\% to 39\%. \wilbur is also within 5\% of the multimodal model, which has access to screenshots during execution, and outperforms it on Allrecipes, ArXiv, Booking.com, ESPN, Cambridge Dictionary, BBC News, and Wolfram.


Comparing against the ablation baselines, we observe that the naive zero-shot baseline is significantly worse than the state of the art, but adding backtracking is enough to come close to the state-of-the-art result. Adding task demonstrations improves to 50\%, showing the value of recalling previous experiences from auto-curriculum.  Finally, the use of the fine-tuned demonstration retrieval model further improves by 3\% overall, highlighting the importance of selecting high-quality task demonstrations.
We additionally discuss the number of LLM calls of the different ablations in Appendix~\ref{sec:wilbur-cost}.

\subsection{Error analysis}
\label{sec:error-analysis}
Further analyzing the cases where \wilbur does not succeed, we observe a few key reasons. These reasons suggest that many of the failures of web agents in practice are engineering-related, not caused by the model, which should inform future research in web agents.

\paragraph{Inability to interact with complex widgets}
A large source of errors for \wilbur was its inability to interact with date selectors, which are used by Google Flights and Booking.com. In these situations, the model tends to get stuck trying to operate these widgets, leading to its eventual failure. Future work should explore creating new built-in functions for the agent to interact with common complex widgets.

\paragraph{Inability to performing actions}
We find that in many cases, even if the agent predicts the correct action, it is unable to perform it on the page. For example, we see issues trying to successfully emulate the Enter key. Additionally, our agent is built to operate only on main DOM of the page, and cannot act inside frames or inside the shadow DOM used by web components. We find this affects the success rates on Apple and GitHub. Future work should investigate a DSL and DOM representation that successfully captures this aspect of web technology, and an agent architecture,

\paragraph{Anti-scraping techniques}
A lot of websites have anti-scraping techniques that detect and block automated agents. This is noticeable on ArXiv, where the agent can be rate-limited and be presented with an empty page. As a result, Wilbur is unable to get past the first step of the execution and fails.
This is a well-known failure mode of automated web agents, with known workarounds such as using proxies and artificially slowing down execution.

\section{Conclusion and Future Work}
We have presented \wilbur, a novel web agent approach that can leverage its own experiences, both successes and failures, in performing the tasks, and automatically improve over time. \wilbur is the first agent that can backtrack to a previous state during execution. This leads to a significantly higher success rate because mistakes become non-fatal even if they are not detected immediately. We also propose a novel in-context-learning approach that combines high-quality positive task demonstrations selected by a fine-tuned model with a large set of positive and negative task demonstrations summarized in succinct instructions.

With our approach, \wilbur achieves a new text-only state-of-the-art result of 53\% on the WebVoyager benchmark, while using the same underlying LLM, showing that there is significant space for improving web agents with better prompting, independently of the base model in use. Our approach is within 5\% of the multimodal state of the art, and in fact surpasses it on Apple, ArXiv, Cambridge Dictionary, BBC News, and Wolfram Alpha, showing that text-only models can be effective at navigating the web, at a fraction of the cost of multi-modal models.

Our error analysis suggests that a large class of errors is in fact caused by the engineering of the agent, and not a failure of the model. These practical issues are a result of the complexity of the web, but we expect they will eventually be overcome, leading to high-accuracy, general-purpose web agents.

Additionally, while \wilbur can learn from executing on similar pages, it still needs to perform expensive model predictions at inference time. In principle, because \wilbur's DSL uses CSS selectors, which generalize across pages with identical structure and different content, successful actions could also be used by future executions without regenerating them. Future work should explore how to directly reuse previously generated skills, both simple and complex, ranging from basic search to filtering and exploring.

\bibliography{custom, acl_anthology}

\begin{thebibliography}{27}
\providecommand{\natexlab}[1]{#1}
\providecommand{\url}[1]{\texttt{#1}}
\expandafter\ifx\csname urlstyle\endcsname\relax
  \providecommand{\doi}[1]{doi: #1}\else
  \providecommand{\doi}{doi: \begingroup \urlstyle{rm}\Url}\fi

\bibitem[Bohra et~al.(2023)Bohra, Verkes, Harutyunyan, Weinberger, and Campagna]{bohra-etal-2023-byoc}
Arth Bohra, Govert Verkes, Artem Harutyunyan, Pascal Weinberger, and Giovanni Campagna.
\newblock {BYOC}: Personalized few-shot classification with co-authored class descriptions.
\newblock In Houda Bouamor, Juan Pino, and Kalika Bali (eds.), \emph{Findings of the Association for Computational Linguistics: EMNLP 2023}, pp.\  13999--14015, Singapore, December 2023. Association for Computational Linguistics.
\newblock \doi{10.18653/v1/2023.findings-emnlp.933}.
\newblock URL \url{https://aclanthology.org/2023.findings-emnlp.933}.

\bibitem[Brown et~al.(2020)Brown, Mann, Ryder, Subbiah, Kaplan, Dhariwal, Neelakantan, Shyam, Sastry, Askell, Agarwal, Herbert-Voss, Krueger, Henighan, Child, Ramesh, Ziegler, Wu, Winter, Hesse, Chen, Sigler, Litwin, Gray, Chess, Clark, Berner, McCandlish, Radford, Sutskever, and Amodei]{NEURIPS2020_1457c0d6}
Tom Brown, Benjamin Mann, Nick Ryder, Melanie Subbiah, Jared~D Kaplan, Prafulla Dhariwal, Arvind Neelakantan, Pranav Shyam, Girish Sastry, Amanda Askell, Sandhini Agarwal, Ariel Herbert-Voss, Gretchen Krueger, Tom Henighan, Rewon Child, Aditya Ramesh, Daniel Ziegler, Jeffrey Wu, Clemens Winter, Chris Hesse, Mark Chen, Eric Sigler, Mateusz Litwin, Scott Gray, Benjamin Chess, Jack Clark, Christopher Berner, Sam McCandlish, Alec Radford, Ilya Sutskever, and Dario Amodei.
\newblock Language models are few-shot learners.
\newblock In H.~Larochelle, M.~Ranzato, R.~Hadsell, M.F. Balcan, and H.~Lin (eds.), \emph{Advances in Neural Information Processing Systems}, volume~33, pp.\  1877--1901. Curran Associates, Inc., 2020.
\newblock URL \url{https://proceedings.neurips.cc/paper_files/paper/2020/file/1457c0d6bfcb4967418bfb8ac142f64a-Paper.pdf}.

\bibitem[Clark et~al.(2003)Clark, Curran, and Osborne]{clark2003bootstrapping}
Stephen Clark, James~R Curran, and Miles Osborne.
\newblock Bootstrapping pos-taggers using unlabelled data.
\newblock In \emph{Proceedings of the seventh conference on Natural language learning at HLT-NAACL 2003}, pp.\  49--55, 2003.

\bibitem[Deng et~al.(2023)Deng, Gu, Zheng, Chen, Stevens, Wang, Sun, and Su]{deng2023mind2web}
Xiang Deng, Yu~Gu, Boyuan Zheng, Shijie Chen, Samuel Stevens, Boshi Wang, Huan Sun, and Yu~Su.
\newblock Mind2web: Towards a generalist agent for the web, 2023.

\bibitem[Furuta et~al.(2023)Furuta, Nachum, Lee, Matsuo, Gu, and Gur]{furuta2023multimodal}
Hiroki Furuta, Ofir Nachum, Kuang-Huei Lee, Yutaka Matsuo, Shixiang~Shane Gu, and Izzeddin Gur.
\newblock Multimodal web navigation with instruction-finetuned foundation models.
\newblock \emph{arXiv preprint arXiv:2305.11854}, 2023.

\bibitem[Gao et~al.(2021)Gao, Fisch, and Chen]{gao-etal-2021-making}
Tianyu Gao, Adam Fisch, and Danqi Chen.
\newblock Making pre-trained language models better few-shot learners.
\newblock In \emph{Proceedings of the 59th Annual Meeting of the Association for Computational Linguistics and the 11th International Joint Conference on Natural Language Processing (Volume 1: Long Papers)}, pp.\  3816--3830, Online, August 2021. Association for Computational Linguistics.
\newblock \doi{10.18653/v1/2021.acl-long.295}.
\newblock URL \url{https://aclanthology.org/2021.acl-long.295}.

\bibitem[Gur et~al.(2022)Gur, Nachum, Miao, Safdari, Huang, Chowdhery, Narang, Fiedel, and Faust]{gur2022understanding}
Izzeddin Gur, Ofir Nachum, Yingjie Miao, Mustafa Safdari, Austin Huang, Aakanksha Chowdhery, Sharan Narang, Noah Fiedel, and Aleksandra Faust.
\newblock Understanding html with large language models.
\newblock \emph{arXiv preprint arXiv:2210.03945}, 2022.

\bibitem[Gur et~al.(2023)Gur, Furuta, Huang, Safdari, Matsuo, Eck, and Faust]{gur2023real}
Izzeddin Gur, Hiroki Furuta, Austin Huang, Mustafa Safdari, Yutaka Matsuo, Douglas Eck, and Aleksandra Faust.
\newblock A real-world webagent with planning, long context understanding, and program synthesis.
\newblock \emph{arXiv preprint arXiv:2307.12856}, 2023.

\bibitem[Haan(2023)]{number-of-websites}
Katherine Haan.
\newblock Top website statistics for 2023.
\newblock \emph{Forbes}, 2023.

\bibitem[He et~al.(2024)He, Yao, Ma, Yu, Dai, Zhang, Lan, and Yu]{he2024webvoyager}
Hongliang He, Wenlin Yao, Kaixin Ma, Wenhao Yu, Yong Dai, Hongming Zhang, Zhenzhong Lan, and Dong Yu.
\newblock Webvoyager: Building an end-to-end web agent with large multimodal models.
\newblock \emph{arXiv preprint arXiv:2401.13919}, 2024.

\bibitem[Hendrycks et~al.(2021)Hendrycks, Burns, Basart, Zou, Mazeika, Song, and Steinhardt]{hendrycks2021measuring}
Dan Hendrycks, Collin Burns, Steven Basart, Andy Zou, Mantas Mazeika, Dawn Song, and Jacob Steinhardt.
\newblock Measuring massive multitask language understanding, 2021.

\bibitem[Kagaya et~al.(2024)Kagaya, Yuan, Lou, Karlekar, Pranata, Kinose, Oguri, Wick, and You]{kagaya2024rap}
Tomoyuki Kagaya, Thong~Jing Yuan, Yuxuan Lou, Jayashree Karlekar, Sugiri Pranata, Akira Kinose, Koki Oguri, Felix Wick, and Yang You.
\newblock Rap: Retrieval-augmented planning with contextual memory for multimodal llm agents, 2024.

\bibitem[Kim et~al.(2024)Kim, Baldi, and McAleer]{kim2024language}
Geunwoo Kim, Pierre Baldi, and Stephen McAleer.
\newblock Language models can solve computer tasks.
\newblock \emph{Advances in Neural Information Processing Systems}, 36, 2024.

\bibitem[Lampinen et~al.(2022)Lampinen, Dasgupta, Chan, Mathewson, Tessler, Creswell, McClelland, Wang, and Hill]{lampinen-etal-2022-language}
Andrew Lampinen, Ishita Dasgupta, Stephanie Chan, Kory Mathewson, Mh~Tessler, Antonia Creswell, James McClelland, Jane Wang, and Felix Hill.
\newblock Can language models learn from explanations in context?
\newblock In \emph{Findings of the Association for Computational Linguistics: EMNLP 2022}, pp.\  537--563, Abu Dhabi, United Arab Emirates, December 2022. Association for Computational Linguistics.
\newblock URL \url{https://aclanthology.org/2022.findings-emnlp.38}.

\bibitem[Liu et~al.(2018)Liu, Guu, Pasupat, Shi, and Liang]{liu2018reinforcement}
Evan~Zheran Liu, Kelvin Guu, Panupong Pasupat, Tianlin Shi, and Percy Liang.
\newblock Reinforcement learning on web interfaces using workflow-guided exploration.
\newblock \emph{arXiv preprint arXiv:1802.08802}, 2018.

\bibitem[Ma et~al.(2023)Ma, Zhang, Wang, Pan, and Yu]{ma2023laser}
Kaixin Ma, Hongming Zhang, Hongwei Wang, Xiaoman Pan, and Dong Yu.
\newblock Laser: Llm agent with state-space exploration for web navigation.
\newblock \emph{arXiv preprint arXiv:2309.08172}, 2023.

\bibitem[McClosky et~al.(2006)McClosky, Charniak, and Johnson]{mcclosky-etal-2006-effective}
David McClosky, Eugene Charniak, and Mark Johnson.
\newblock Effective self-training for parsing.
\newblock In \emph{Proceedings of the Human Language Technology Conference of the {NAACL}, Main Conference}, pp.\  152--159, New York City, USA, June 2006. Association for Computational Linguistics.
\newblock URL \url{https://aclanthology.org/N06-1020}.

\bibitem[Min et~al.(2022)Min, Lyu, Holtzman, Artetxe, Lewis, Hajishirzi, and Zettlemoyer]{min2022rethinking}
Sewon Min, Xinxi Lyu, Ari Holtzman, Mikel Artetxe, Mike Lewis, Hannaneh Hajishirzi, and Luke Zettlemoyer.
\newblock Rethinking the role of demonstrations: What makes in-context learning work?
\newblock \emph{arXiv preprint arXiv:2202.12837}, 2022.

\bibitem[OpenAI(2023)]{openai2023gpt4}
OpenAI.
\newblock Gpt-4 technical report, 2023.

\bibitem[Shi et~al.(2017)Shi, Karpathy, Fan, Hernandez, and Liang]{shi2017world}
Tianlin Shi, Andrej Karpathy, Linxi Fan, Jonathan Hernandez, and Percy Liang.
\newblock World of bits: An open-domain platform for web-based agents.
\newblock In \emph{International Conference on Machine Learning}, pp.\  3135--3144. PMLR, 2017.

\bibitem[Sridhar et~al.(2023)Sridhar, Lo, Xu, Zhu, and Zhou]{sridhar2023hierarchical}
Abishek Sridhar, Robert Lo, Frank~F Xu, Hao Zhu, and Shuyan Zhou.
\newblock Hierarchical prompting assists large language model on web navigation.
\newblock \emph{arXiv preprint arXiv:2305.14257}, 2023.

\bibitem[Sun et~al.(2024)Sun, Zhuang, Kong, Dai, and Zhang]{sun2024adaplanner}
Haotian Sun, Yuchen Zhuang, Lingkai Kong, Bo~Dai, and Chao Zhang.
\newblock Adaplanner: Adaptive planning from feedback with language models.
\newblock \emph{Advances in Neural Information Processing Systems}, 36, 2024.

\bibitem[Wang et~al.(2023)Wang, Xie, Jiang, Mandlekar, Xiao, Zhu, Fan, and Anandkumar]{wang2023voyager}
Guanzhi Wang, Yuqi Xie, Yunfan Jiang, Ajay Mandlekar, Chaowei Xiao, Yuke Zhu, Linxi Fan, and Anima Anandkumar.
\newblock Voyager: An open-ended embodied agent with large language models.
\newblock \emph{arXiv preprint arXiv: Arxiv-2305.16291}, 2023.

\bibitem[Xu et~al.(2021)Xu, Masling, Du, Campagna, Heck, Landay, and Lam]{xu2021grounding}
Nancy Xu, Sam Masling, Michael Du, Giovanni Campagna, Larry Heck, James Landay, and Monica~S Lam.
\newblock Grounding open-domain instructions to automate web support tasks, 2021.

\bibitem[Yao et~al.(2022{\natexlab{a}})Yao, Chen, Yang, and Narasimhan]{NEURIPS2022_82ad13ec}
Shunyu Yao, Howard Chen, John Yang, and Karthik Narasimhan.
\newblock Webshop: Towards scalable real-world web interaction with grounded language agents.
\newblock In S.~Koyejo, S.~Mohamed, A.~Agarwal, D.~Belgrave, K.~Cho, and A.~Oh (eds.), \emph{Advances in Neural Information Processing Systems}, volume~35, pp.\  20744--20757. Curran Associates, Inc., 2022{\natexlab{a}}.
\newblock URL \url{https://proceedings.neurips.cc/paper_files/paper/2022/file/82ad13ec01f9fe44c01cb91814fd7b8c-Paper-Conference.pdf}.

\bibitem[Yao et~al.(2022{\natexlab{b}})Yao, Zhao, Yu, Du, Shafran, Narasimhan, and Cao]{yao2022react}
Shunyu Yao, Jeffrey Zhao, Dian Yu, Nan Du, Izhak Shafran, Karthik Narasimhan, and Yuan Cao.
\newblock React: Synergizing reasoning and acting in language models.
\newblock \emph{arXiv preprint arXiv:2210.03629}, 2022{\natexlab{b}}.

\bibitem[Zheng et~al.(2024)Zheng, Gou, Kil, Sun, and Su]{zheng2024gpt4vision}
Boyuan Zheng, Boyu Gou, Jihyung Kil, Huan Sun, and Yu~Su.
\newblock Gpt-4v(ision) is a generalist web agent, if grounded, 2024.

\end{thebibliography}
\bibliographystyle{colm2024_conference}

\newpage
\appendix

\section{DOM Formatting}
\label{sec:dom-formatting}

To represent the DOM, we include all leaves of interactive and structural elements (headings, links, paragraphs). We do not include formatting elements. For each element, we include element index, tag name, accessiblity properties (role, alt, ARIA label), content, and links. An example snippet of formatted DOM is included in Fig.~\ref{fig:dom-example}.

\begin{figure}[h!]
\begin{verbatim}
<0: [button] ariaLabel: Collapse side panel/>
<1: [button] ariaLabel: Collapse side panel/>
<2: [label] text: “Search Google Maps”/>
<3: [label] text: Search Google Maps/>
<4: [input] required: True, name: q, type: text/>
<5: [button] ariaLabel: Search/>
<6: [button] ariaLabel: Directions/>
<7: [button] ariaLabel: Collapse side panel/>
<8: [button] ariaLabel: Collapse side panel/>
<9: [button] ariaLabel: Menu/>
<10: [button] text: “Saved”/>
<11: [button] text: “Recents”/>
<12: [button] ariaLabel: Collapse side panel/>
<13: [button] ariaLabel: Collapse side panel/>
<14: [a] ariaLabel: Sign in, text: “Sign in”/>
<15: [button] ariaLabel: Show Your Location/>
<16: [label] text: “Show Your Location”/>
<17: [label] text: Show Your Location/>
<18: [button] text: “Update”/>
<19: [button] text: “Learn more”/>
<20: [button] ariaLabel: Zoom in/>
<21: [label] text: “Zoom”/>
<22: [label] text: Zoom/>
<23: [button] text: “Show slider”/>
<24: [button] text: “Hide slider”/>
<25: [button] ariaLabel: Zoom out/>
<26: [button] ariaLabel: Show Street View coverage/>
<27: [button] ariaLabel: Show imagery/>
<28: [button] ariaLabel: Collapse side panel/>
<29: [button] ariaLabel: Zoom in/>
<30: [button] ariaLabel: Zoom out/>
<31: [span] text: ""/>
<32: [label] text: “Layers”/>
<33: [label] text: Layers/>
<34: [h2] text: “Map details”/>
<35: [button] ariaLabel: Close, text: “”/>
<36: [button] text: “Transit”/>
<37: [button] text: “Traffic”/>
<38: [button] text: “Biking”/>
<39: [button] text: “Terrain”/>
<40: [button] text: “Street View”/>
<41: [button] text: “Wildfires”/>
<42: [button] text: “Air Quality”/>
<43: [h2] text: “Map tools”/>
<44: [button] text: “Travel time”/>
<45: [button] text: “Measure”/>
<46: [h2] text: “Map type”/>
<47: [button] text: “Default”/>
<48: [button] text: “Satellite”/>
<49: [button] text: “Globe view”/>
<50: [button] text: “Labels”/>
<51: [span] text: Map data ©2024 Google/>
<52: [button] text: “United States”/>
<53: [button] text: “Terms”/>
<54: [button] text: “Privacy”/>
<55: [button] text: “Send Product Feedback”/>
<56: [button] text: “2000 ft”>
\end{verbatim}
\caption{Example of formatted DOM from Google Maps}
\label{fig:dom-example}
\end{figure}

\section{DSL Definition}
\label{sec:dsl-definition}

The \wilbur actor predicts code in a DSL syntactically similar to Python. The DSL does not support control constructs, and only supports assignments and function call statements. The agent can predict more than one function call in one step, and they are executed sequentially. The agent predicts element references by numeric index, which are converted to CSS selectors prior to execution.

The DSL has access to the following builtin functions:

{
\small
\begin{tabular}{lp{9cm}}
\toprule
\texttt{save\_text}(\textit{element}, \textit{key}) & Extract the content of the given element into the agent's temporary database. \textit{key} is the description of the saved text. \\
\hline
\texttt{save\_link}(\textit{element}, \textit{key}) & Extract the link URL from the given anchor element into the agent's temporary database. \\
\hline
\texttt{save\_list}($\textit{element}_1$, $\textit{element}_2$) & Given two elements inside a list, compute a CSS selector that refers to all elements in the list, then enter \textit{list mode}. While in list mode, further DSL calls are repeated for each list element, using CSS selectors scoped to the list element.\\
\hline
\texttt{type\_input}(\textit{element}, \textit{value}) & Set the given element to the specified value. \\
\hline
\texttt{press\_enter}(\textit{element}) & Focus the given element and press the Enter key. \\
\hline
\texttt{click}(\textit{element}) & Click on the given element. \\
\hline
\texttt{go\_back}() & Navigate to the previous URL. \\
\bottomrule
\end{tabular}
}

\section{Training Algorithm}

\begin{algorithm}[h]
\caption{Training \wilbur}
\label{algo:wilbur_training}
\small
\begin{algorithmic}[1]
\Require Websites $W$
\State $D_\tau, D_a \gets \emptyset$ \Comment{Knowledge and action demonstration vectors}
\State Initialize auto-curriculum phase
\For{each website $w \in W$}
    \State $o_0 \gets \text{initial observation state in } w$
    \State $G \gets \text{GenerateGoals}(w)$ \Comment{Generate multiple plausible goals}
    \For{each goal $g \in G$}
        \State $\tau, \textit{answer} \gets \text{\wilbur}(g,o_0)$ \Comment{Perform actions towards goal}
        \State $v_\tau \gets \text{ExecutionSelfEvaluation}(\tau, g, text)$
        \State $\text{UpdateDemonstrations}(D_g, D_a, g, v_\tau)$
    \EndFor
    \State $G' \gets \text{GenerateFollowupGoals}(G, w)$ \Comment{Condition on previous goals}
    \For{each goal $g' \in G'$}
        \State $\tau, text \gets \text{\wilbur}(g,o_0)$ \Comment{Perform actions towards goal}
        \State $\hat{r} \gets \text{ExecutionSelfEvaluation}(\tau, text)$
        \State $\text{UpdateDemonstrations}(D_\tau, D_a, g', v_\tau)$
    \EndFor
\EndFor
\State $\text{TrainKnowledgeModel}(D_g, D_a)$ \Comment{Finetune model with demonstrations}
\end{algorithmic}
\end{algorithm}

\section{Computation Cost of \wilbur}
\label{sec:wilbur-cost}

In Table~\ref{table:token-usage} we show the number of actor LLM calls used by \wilbur and the different ablated baselines. We see that while adding in-context learning examples decreases the number of steps required by the agent, the added synthesis step leads to a sharp increase again. \wilbur, however, utilizing the demonstrating ranking model, is able to filter out bad fewshot examples to lower the average steps to success.

\begin{table}[t]
\caption{Average number of actor LLM calls on successful executions on the WebVoyager benchmark}
\label{table:token-usage}
\small
\centering
\begin{tabular}{lr}
\toprule
 & \bf \# calls \\
\midrule
Zero-shot & 6.72 \\
$+$ Backtrack & 6.43 \\
$+$ Demonstrations & \textbf{4.34} \\
$+$ Synthesis & 7.19 \\
\wilbur & 5.33 \\
\bottomrule
\end{tabular}

\end{table}

\end{document}